\documentclass[11pt,a4paper,logo]{lumia}
\usepackage[round,authoryear]{natbib}
\setheadertext{LUMIA Lab}

\usepackage{amsmath,amsfonts,bm}

\def\eqref#1{equation~\ref{#1}}
\def\1{\bm{1}}

\DeclareMathAlphabet{\mathsfit}{\encodingdefault}{\sfdefault}{m}{sl}
\SetMathAlphabet{\mathsfit}{bold}{\encodingdefault}{\sfdefault}{bx}{n}

\newcommand{\softplus}{\zeta}

\usepackage{xurl}
\usepackage{amsmath}
\usepackage{amssymb}
\usepackage{mathtools}
\usepackage{booktabs}
\usepackage{graphicx}
\usepackage{tabularx}
\usepackage{multirow}
\usepackage{xcolor}
\usepackage{algorithm}
\usepackage{algorithmic}
\usepackage{float}
\usepackage{flafter}
\usepackage{placeins}
\usepackage{needspace}
\usepackage{array}
\usepackage{hyperref}
\definecolor{bigaired}{RGB}{156,0,0}
\definecolor{color5}{HTML}{006795}
\hypersetup{colorlinks=true,linkcolor=bigaired,citecolor=color5,urlcolor=color5}

\title{StateLoop: Persistent State\\for Looped Transformers}

\author{%
Feichen~Song$^{1}$, Boyi~Zeng$^{1}$, He~Li$^{1}$, Shixiang~Song$^{1}$, Yiqin~Hao$^{1}$,
Zitong~Wang$^{2}$, Yizhou~Yang$^{1}$, Yifan~Liu$^{1}$, Ziwei~He$^{3}$, Zhouhan~Lin$^{1}$\\
$^{1}$Shanghai Jiao Tong University\\
$^{2}$Sun Yat-sen University \quad $^{3}$Shanghai Innovation Institute
}

\newcommand{\method}{StateLoop}
\newcommand{\methodattn}{\method{} + Attn Mixer}
\newcommand{\methodmamba}{\method{} + Mamba-2 Mixer}
\newcommand{\RMSNorm}{\operatorname{RMSNorm}}

\newcommand{\MambaTwoBlock}{\operatorname{Mamba2Block}}
\newcommand{\CausalAttn}{\operatorname{CausalAttention}}
\renewcommand{\softplus}{\operatorname{softplus}}
\newcommand{\LMHead}{\operatorname{LMHead}}
\renewcommand{\algorithmicrequire}{\textbf{Require}}

\newcommand{\methodattnshort}{\method{} + Attn}
\newcommand{\methodmambashort}{\method{} + Mamba-2}
\newcommand{\tablecaptiongap}{\vspace{4pt}}

\begin{document}

\begin{abstract}
Looped Transformers offer a parameter-efficient approach to language modeling by repeatedly applying a shared backbone. Vanilla looped Transformers use each backbone output directly as the input to the next iteration. These outputs may contain information useful several iterations later, but retaining it across the loop depends entirely on the single hidden state. We introduce \textbf{\method{}}, which maintains one additional persistent state per token to retain information across iterations. At each iteration, a learned gate controls how much of the previous state is retained and how much information from the backbone output is incorporated. The updated state is projected back and added to the backbone output for subsequent iterations and final prediction. This design reduces Pile perplexity and improves average accuracy across eight zero-shot tasks over Vanilla Looped. Its perplexity advantage over Vanilla Looped grows with trained loop depth, consistent with persistent state becoming increasingly useful as recurrent computation deepens.
\end{abstract}
\maketitle

\section{Introduction}

Looped Transformers increase computation by repeatedly applying a shared backbone \citep{dehghani2019universal}. \citet{saunshi2025latent} show that looped models improve language modeling over non-looped models with the same number of parameters. Ouro demonstrates this approach at larger scale, with 1.4B and 2.6B looped models achieving performance comparable to larger standard Transformers across downstream benchmarks \citep{zhu2025ouro}.

The design of connections between iterations is a central architectural consideration in looped Transformers. Vanilla looped Transformers use each backbone output directly as the input to the next iteration. As \citet{pagliardini2024denseformer} show, DenseFormer achieves lower perplexity when later layers can access much earlier outputs than when access is restricted to recent outputs. This finding motivates the hypothesis that information produced in one iteration may likewise remain useful several iterations later. In vanilla looped Transformers, retaining such information across iterations relies entirely on the same hidden state that also supplies the input for subsequent backbone computation.

We therefore propose \method{}, which maintains an additional persistent state for each token. The state is carried across iterations and updated by combining its previous value with a projection of the backbone output. A gate determined by the same output controls how much of the previous state is retained and how much new information is written. An optional state mixer uses causal attention or Mamba-2 to exchange information across token positions before readout. The updated state is projected back and added to the backbone output, making the retained information available to later iterations and final prediction.

We evaluate \method{} at 410M and 1.4B scales to assess both its performance and the contribution of the persistent state. Compared with Vanilla Looped, \method{} reduces Pile perplexity by 0.18 and improves average accuracy across eight zero-shot tasks by 0.97 percentage points at 1.4B. Adding causal attention and Mamba-2 state mixers further increases the accuracy gains over Vanilla Looped to 1.43 and 1.44 percentage points, respectively. We further analyze the role of the persistent state. Resetting the state at every iteration increases perplexity, supporting the value of carrying information across iterations. Moreover, \method{}'s advantage over Vanilla Looped grows with trained loop depth, suggesting that persistent state becomes increasingly useful as recurrent computation deepens.

\section{Related Work}

\paragraph{Looped Transformers and recurrent depth.}
The Universal Transformer introduced recurrent computation across depth by repeatedly applying a shared block \citep{dehghani2019universal}. Subsequent work shows that looped Transformers can learn iterative algorithms for in-context learning \citep{yang2023learning}. Ouro extends this approach to large-scale language-model pretraining with learned depth allocation \citep{zhu2025ouro}.

\paragraph{Connections between iterations.}
Recent looped language models differ in how they carry information across iterations and construct the next backbone input. Parcae stabilizes the recurrent hidden state by constraining the spectral norm of its state-transition matrix \citep{prairie2026parcae}. PonderLM feeds back a weighted sum of token embeddings, with weights given by the predicted token probabilities \citep{zeng2025ponder}. Hyperloop carries multiple residual streams across iterations and uses loop-specific hyper-connections to read the backbone input from these streams and write the output back \citep{zeitoun2026hyperloop}. These methods modify or enrich the recurrent stream itself, whereas \method{} retains the direct backbone stream and introduces a separate persistent state.

\paragraph{Persistent state and memory.}
LSTMs, GRUs, and selective state-space models update recurrent state along a sequence \citep{hochreiter1997lstm,cho2014gru,gu2023mamba}, while Recurrent Memory Transformers carry memory tokens between sequence segments \citep{bulatov2022rmt}. In looped architectures, Adaptive Loops and Memory augments repeated computation with gated access to static learned memory banks \citep{frey2026adaptive}, while Universal Transformers Need Memory studies dynamically updated memory tokens as a computational scratchpad for recursive reasoning \citep{sapunov2026memory}. MeSH externalizes recurrent state management into a multi-slot buffer with iteration-specific read and write routers \citep{yu2026mesh}. \method{} explores a simpler design point: it maintains a single continuously updated state per token, retains the main recurrent hidden stream as a direct path, and shares all state-update and feedback parameters at every iteration.

\section{Method}
\label{sec:method}

\begin{figure}[!t]
    \centering
    \includegraphics[width=\linewidth,trim=0 7bp 0 7bp,clip]{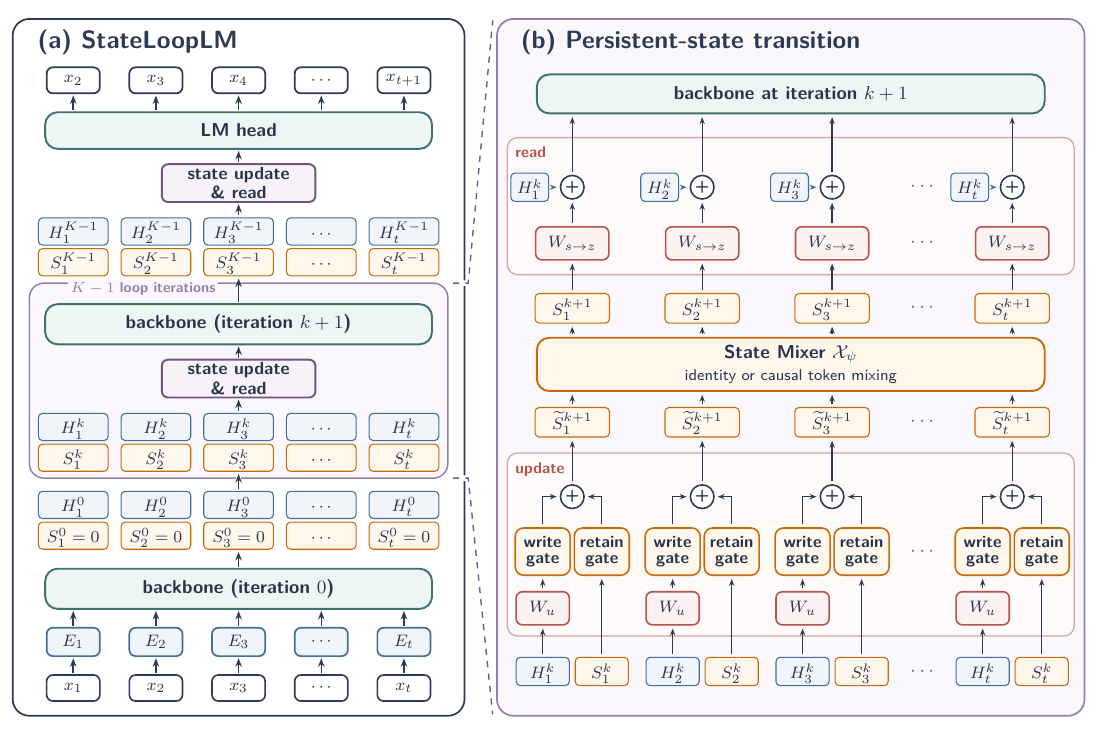}
    \caption{\textbf{\method{} architecture.} (a) The looped model with a persistent state per token. (b) State update, optional token mixing, and readout. \(W_{s\to z}\) feeds the next iteration, whereas \(W_{s\to out}\) supplies the final prediction.}
    \label{fig:overview}
\end{figure}

\method{} adds a persistent state to each token. At each iteration, a gated update incorporates the backbone output into the state, which is then read to form the next backbone input or the final prediction. Figure~\ref{fig:overview} shows this computation.

\subsection{Backbone Feedback and Persistent State}
\label{sec:coupled_streams}

Let \(x=(x_1,\ldots,x_n)\) be a token sequence and \(T\) the shared Transformer backbone, with backbone width \(d\) and state width \(d_s\). We use \(K\geq1\) loop iterations in total, indexed by \mbox{\(k=0,\ldots,K-1\)}. Initialization is
\begin{equation}
E=\sqrt d\,\operatorname{Embed}(x),
\qquad S^0=0,
\qquad H^0=T(E).
\label{eq:init}
\end{equation}
Here \(H^k\in\mathbb{R}^{n\times d}\) is the backbone output at iteration \(k\), and \(S^k\in\mathbb{R}^{n\times d_s}\) is the state after \(k\) updates, before incorporating \(H^k\). Updating it with \(H^k\) produces \(S^{k+1}\). Linear projections act on each token's feature vector.

At iteration \(k=0,\ldots,K-1\), the state branch receives the backbone output normalized with RMSNorm \citep{zhang2019rmsnorm},
\begin{equation}
U^k=\RMSNorm_h(H^k).
\label{eq:backbone_step}
\end{equation}
The state update \(\mathcal{V}_{\phi}\) combines \(S^k\) and \(U^k\) independently at each token. The operator \(\mathcal{X}_{\psi}\) then applies identity or causal token mixing. The resulting state is normalized for readout, giving
\begin{equation}
\widetilde S^{k+1}=\mathcal{V}_{\phi}(S^k,U^k),
\qquad S^{k+1}=\mathcal{X}_{\psi}(\widetilde S^{k+1}),
\qquad N^{k+1}=\RMSNorm_s(S^{k+1}).
\label{eq:state_step}
\end{equation}
Section~\ref{sec:state_transition} defines the state update and the mixing options.

For \(k=0,\ldots,K-2\), the backbone output \(H^k\) and the projected state read form the input \(Z^{k+1}\in\mathbb{R}^{n\times d}\) to the next iteration. The backbone then computes its next output from this input,
\begin{equation}
\boxed{\begin{aligned}
Z^{k+1}&=H^k+N^{k+1}W_{s\rightarrow z}^{\top},\\
H^{k+1}&=T(Z^{k+1}).
\end{aligned}}
\label{eq:state_feedback}
\end{equation}
The projection \(W_{s\rightarrow z}\in\mathbb{R}^{d\times d_s}\) maps the state read to the backbone width. The backbone, state update, feedback projection, and optional token mixer share their parameters across iterations.

\subsection{State Update}
\label{sec:state_transition}

\subsubsection{Selective SSM Update}
\label{sec:selective_update}

The selective-SSM recurrence operates along the loop axis independently for each token, while \(\mathcal{X}_{\psi}\) provides optional mixing along the token axis. A learned projection maps the normalized backbone output to the state width, giving
\begin{equation}
P^k=U^kW_u^{\top},
\qquad W_u\in\mathbb{R}^{d_s\times d}.
\end{equation}
The normalized backbone output \(U^k\) determines an input-dependent step size, following selective SSMs \citep{gu2023mamba}. We use the update
\begin{align}
\Delta^k&=\softplus\!\left(U^kW_\Delta^{\top}+b_\Delta\right),
&\lambda&=\exp(A_{\log}),\nonumber\\
\widetilde S^{k+1}
&=\exp(-\lambda\odot\Delta^k)\odot S^k
+\bigl(1-\exp(-\lambda\odot\Delta^k)\bigr)\odot P^k.
\label{eq:ssm_vertical}
\end{align}
Here \(W_\Delta\in\mathbb{R}^{d_s\times d}\), \(b_\Delta,A_{\log}\in\mathbb{R}^{d_s}\), and \(\odot\) denotes elementwise multiplication. The positive rate \(\lambda\) is learned per channel. The retention gate \(g^k=\exp(-\lambda\odot\Delta^k)\) weights the previous state, while the write gate \(\eta^k=1-g^k\) weights the projected output \(P^k\). Because \(\Delta^k\) depends on \(U^k\), these weights can vary with the token and the iteration.

We initialize $W_\Delta = 0$ and
$b_\Delta = \log(e^{0.3}-1)$, so that $\Delta = 0.3$ initially,
and initialize $\lambda = \exp(A_{\log})$ logarithmically from
$0.01$ to $4$ across channels. These parameters are shared across
iterations and remain trainable. The resulting initial write
rate is $\eta_c = 1 - \exp(-0.3\lambda_c)$.

\subsubsection{Optional Token-Axis Mixing}
\label{sec:causal_mixer}

\method{} uses identity mixing, so the updated state is normalized and read out directly, with
\begin{equation}
S^{k+1}=\mathcal{X}_{\mathrm{id}}(\widetilde S^{k+1})=\widetilde S^{k+1}.
\label{eq:identity_mixer}
\end{equation}
Adding causal attention or Mamba-2 gives \methodattn{} or \methodmamba{}, respectively, allowing updated states to exchange information across token positions. Both preserve the shape \(n\times d_s\) and compute the output at position \(t\) from the updated-state prefix \(\widetilde S^{k+1}_{\leq t}\).

The causal-attention mixer uses self-attention \citep{vaswani2017attention} with pre-normalization and a residual connection, giving
\begin{equation}
S^{k+1}=\widetilde S^{k+1}
+\CausalAttn_{\psi}\!\left(\RMSNorm_m(\widetilde S^{k+1})\right).
\label{eq:attention_mixer}
\end{equation}

The Mamba-2 mixer applies a residual Mamba-2/SSD block along the token axis \citep{dao2024transformers} to obtain
\begin{equation}
S^{k+1}=\MambaTwoBlock_{\psi}(\widetilde S^{k+1}).
\label{eq:state_mixer}
\end{equation}
\subsection{Final Prediction and Training}
\label{sec:prediction}

At the final iteration \(k=K-1\), \(H^{K-1}\) updates \(S^{K-1}\) to \(S^K\). A separate readout projection \(W_{s\rightarrow out}\in\mathbb{R}^{d\times d_s}\) combines the normalized state with \(H^{K-1}\) to produce
\begin{equation}
\boxed{
Y
=H^{K-1}
+\RMSNorm_s(S^K)W_{s\rightarrow out}^{\top}
=H^{K-1}+N^KW_{s\rightarrow out}^{\top}.
}
\label{eq:output}
\end{equation}

The LM head maps \(Y\) to vocabulary logits. All parameters are trained end-to-end by minimizing the standard autoregressive loss,
\begin{align}
p_\theta(\,\cdot \mid x_{\leq t})
&=\operatorname{softmax}\bigl(\LMHead(Y_t)\bigr),\\
\mathcal{L}_{\mathrm{LM}}
&=-\frac{1}{n-1}\sum_{t=1}^{n-1}
\log p_\theta(x_{t+1}\mid x_{\leq t}).
\label{eq:lm_loss}
\end{align}

\section{Experiments}
\label{sec:experiments}

We evaluate language modeling and downstream accuracy, examine performance across trained loop depths, and measure autoregressive decoding efficiency. Section~\ref{sec:ablation_analysis} examines the design and behavior of the persistent state.

\subsection{Experimental Setup}
\label{sec:setup}

\paragraph{Training configuration.}
All models are trained from scratch on the Pile \citep{gao2020pile} with a
LLaMA-style backbone \citep{touvron2023llama}. All 1.4B comparisons use the
same GPT-NeoX tokenizer, data pipeline, 30B-token training budget, batch size 256, sequence length 2,048, and cosine-decay schedule with peak learning rate
\(2\times10^{-4}\). The shared Transformer backbone contains 24 layers,
with hidden width \(d=2048\), SwiGLU intermediate width 5504, and 32
attention heads of dimension 64. \method{} uses persistent-state width
\(d_s=2048\), and the full-backbone recurrent models use \(K=4\)
iterations of the shared backbone. The 410M studies use 25B training
tokens and a peak learning rate of \(3\times10^{-4}\), with the same
batch size, sequence length, and cosine-decay schedule. Training budgets
count original language-model tokens.

\paragraph{State-branch configuration.}
Both token-axis mixers consist of a single residual module whose parameters are shared across iterations. In the 1.4B model, the attention mixer uses pre-normalized causal self-attention with rotary position embeddings (RoPE), 32 heads of dimension 64, and no FFN. The Mamba-2 mixer uses an expansion factor of 2, an internal SSM state dimension of 128, a convolution width of 4, a head dimension of 64, and 8 groups, without additional positional embeddings.

\paragraph{Comparison models.}
All baselines are implemented in the same codebase. Vanilla Looped
\citep{dehghani2019universal} and PonderLM \citep{zeng2025ponder} apply the
full backbone for four iterations. Vanilla Looped + FFN increases the SwiGLU
intermediate width from 5,504 to 5,818, yielding approximately the same
parameter count as \methodmamba{}.
Our Hyperloop \citep{zeitoun2026hyperloop}, MeSH \citep{yu2026mesh}, and
Parcae \citep{prairie2026parcae} baselines retain the partial-loop structures
used in their original studies. Hyperloop follows its original begin--middle--end
structure, with only the middle block applied recurrently. For MeSH and Parcae,
we use eight layers each for the prelude, recurrent core, and coda, with ten
core repetitions, matching the 96 layer applications of four \method{}
iterations, each with 24 layers.
Pause Token inserts three pause tokens after each input token to provide
additional computation before predicting the next token \citep{goyal2023pause}.

\paragraph{Evaluation metrics.}
We report perplexity (PPL) on Pile validation data, LAMBADA-OpenAI (LAMBADA-OAI), and WikiText-2. PPL is the exponential of mean negative log-likelihood (NLL). Pile and LAMBADA-OAI use token PPL, while WikiText-2 uses word PPL \citep{paperno2016lambada,radford2019language,merity2017pointer}. Downstream evaluation covers LAMBADA-OAI, ARC-Easy, ARC-Challenge, WinoGrande, PIQA, HellaSwag, SciQ, and RACE-high (RACE in the tables) under both zero-shot and five-shot settings \citep{clark2018arc,sakaguchi2020winogrande,bisk2020piqa,zellers2019hellaswag,welbl2017sciq,lai2017race}. We report task accuracy and Avg-8, the unweighted average across these eight tasks.

\subsection{Main Results}
\label{sec:main_results}

Tables~\ref{tab:main_ppl} and~\ref{tab:main_downstream} compare three \method{} variants with the baselines under the common 1.4B training setup. The variants share the selective-SSM update and feedback rule and apply identity, causal attention, or Mamba-2 to the updated state.

\begin{table}[!htbp]
\caption{\textbf{Language-modeling PPL at 1.4B (30B training tokens).} Full-backbone models use \(K=4\), with MeSH and Parcae matched at 96 layer applications. LAMBADA-OpenAI uses zero-shot token PPL and WikiText-2 uses word PPL.}
\label{tab:main_ppl}
\centering
\tablecaptiongap
\small
\setlength{\tabcolsep}{4.0pt}
\renewcommand{\arraystretch}{1.00}
\begin{tabular*}{0.96\linewidth}{@{\extracolsep{\fill}}lccccc@{}}
\toprule
& \multicolumn{2}{c}{Parameters} & \multicolumn{3}{c}{Perplexity $\downarrow$} \\
\cmidrule(lr){2-3}\cmidrule(lr){4-6}
Model & Params (M) & \shortstack{vs. Vanilla\\Looped} & Pile & \shortstack{LAMBADA\\OAI} & WikiText-2 \\
\midrule
Vanilla Looped & 1420.92 & 0.00\% & 7.79 & 6.94 & 16.50 \\
Vanilla Looped + FFN & 1467.22 & +3.26\% & 7.75 & 7.00 & 16.34 \\
Hyperloop & 1421.32 & +0.03\% & 7.77 & 6.84 & 16.66 \\
MeSH & 1421.51 & +0.04\% & 7.74 & 6.77 & 16.43 \\
PonderLM & 1420.92 & 0.00\% & 7.85 & 7.31 & 16.72 \\
Parcae & 1429.32 & +0.59\% & 7.87 & 7.38 & 16.83 \\
Pause Token & 1420.92 & 0.00\% & 8.20 & 7.74 & 17.87 \\
\midrule
\method{} & 1437.71 & +1.18\% & \underline{7.61} & \underline{6.71} & 15.96 \\
\methodattnshort{} & 1454.49 & +2.36\% & \textbf{7.60} & \textbf{6.64} & \textbf{15.93} \\
\methodmambashort{} & 1467.23 & +3.26\% & \textbf{7.60} & 6.88 & \underline{15.95} \\
\bottomrule
\end{tabular*}
\end{table}

\begin{table}[!htbp]
\caption{\textbf{Zero-shot and five-shot accuracy (\%) at 1.4B after 30B training tokens.} Full-backbone models use \(K=4\); MeSH and Parcae match their 96 layer applications. Avg-8 is the unweighted eight-task mean. LAMBADA-OAI denotes LAMBADA-OpenAI.}
\label{tab:main_downstream}
\centering
\tablecaptiongap
\small
\setlength{\tabcolsep}{1.5pt}
\renewcommand{\arraystretch}{1.00}
\begin{tabular*}{\linewidth}{@{\extracolsep{\fill}}lccccccccc@{}}
\toprule
Model & \shortstack{LAMBADA\\OAI} & ARC-E & ARC-C & WinoG. & PIQA & Hella. & SciQ & RACE & Avg-8 \\
\midrule
\multicolumn{10}{@{}l}{\textit{Zero-shot}} \\
Vanilla Looped & 58.24 & 52.57 & 28.16 & 54.85 & 70.02 & 50.39 & 77.90 & 34.35 & 53.31 \\
Vanilla Looped + FFN & 59.05 & 50.93 & 26.71 & 56.91 & 69.42 & 50.50 & 75.60 & 33.68 & 52.85 \\
Hyperloop & \textbf{61.28} & 51.05 & 27.56 & 55.72 & \underline{70.46} & 49.69 & \underline{80.70} & 34.16 & 53.83 \\
MeSH & \underline{60.94} & 50.88 & 26.62 & \textbf{59.67} & 69.70 & 50.52 & 78.10 & 34.55 & 53.87 \\
PonderLM & 58.26 & 50.59 & \underline{28.58} & 55.41 & 68.93 & 48.96 & 78.00 & 33.68 & 52.80 \\
Parcae & 57.40 & 50.04 & 26.71 & 53.83 & \underline{70.46} & 49.34 & 78.00 & 34.45 & 52.53 \\
Pause Token & 57.05 & 49.45 & 25.77 & 53.12 & 68.28 & 45.42 & 75.00 & 32.25 & 50.79 \\
\addlinespace[1pt]
\method{} & 60.68 & 52.27 & 28.33 & 56.43 & 70.13 & 51.19 & 80.60 & \underline{34.64} & 54.28 \\
\methodattnshort{} & 59.91 & \underline{53.79} & \textbf{29.35} & 56.43 & 70.40 & \underline{51.61} & \textbf{81.50} & \textbf{34.93} & \underline{54.74} \\
\methodmambashort{} & 58.99 & \textbf{55.39} & 27.90 & \underline{59.04} & \textbf{71.06} & \textbf{52.14} & 79.20 & 34.26 & \textbf{54.75} \\
\midrule
\multicolumn{10}{@{}l}{\textit{Five-shot}} \\
Vanilla Looped & 51.64 & 60.23 & 29.69 & 55.17 & 69.53 & 50.02 & 90.90 & 33.78 & 55.12 \\
Vanilla Looped + FFN & 53.15 & 60.06 & 29.52 & 57.06 & 70.62 & 50.85 & 91.40 & 31.67 & 55.54 \\
Hyperloop & \textbf{55.81} & 58.84 & \underline{31.66} & 58.25 & 69.80 & 50.45 & 92.00 & 34.35 & 56.39 \\
MeSH & \underline{54.94} & 59.81 & 29.86 & 58.41 & 69.21 & 51.55 & \textbf{92.70} & \textbf{35.50} & 56.50 \\
PonderLM & 52.53 & 57.95 & 29.35 & 55.33 & 68.93 & 49.36 & 91.80 & 33.30 & 54.82 \\
Parcae & 50.94 & 59.51 & 30.63 & 54.93 & 70.08 & 49.59 & 91.50 & 33.01 & 55.02 \\
Pause Token & 50.79 & 55.98 & 28.16 & 53.28 & 68.50 & 45.45 & 88.60 & 32.44 & 52.90 \\
\addlinespace[1pt]
\method{} & 53.89 & \underline{61.49} & 30.55 & 58.09 & 70.02 & \underline{52.24} & 92.20 & 34.35 & 56.60 \\
\methodattnshort{} & 54.71 & 60.86 & \textbf{31.91} & \textbf{59.04} & \underline{71.11} & \textbf{52.62} & 91.90 & \underline{35.02} & \textbf{57.15} \\
\methodmambashort{} & 53.81 & \textbf{62.58} & 30.80 & \underline{58.80} & \textbf{71.71} & 52.18 & \underline{92.40} & 34.74 & \underline{57.13} \\
\bottomrule
\end{tabular*}
\end{table}

Relative to Vanilla Looped, \method{} lowers perplexity on all three datasets and improves average downstream accuracy under both prompting settings (Tables~\ref{tab:main_ppl} and~\ref{tab:main_downstream}). Both token mixers further improve Avg-8, with similar scores in each setting. 

In this parameter-control comparison, allocating the additional parameters to the state branch yields better performance than widening the FFN. \method{} outperforms Vanilla Looped + FFN in Pile PPL and both downstream averages with fewer parameters, while \methodmamba{} does so at a comparable parameter count (Tables~\ref{tab:main_ppl} and~\ref{tab:main_downstream}).

\subsection{Loop Depth}
\label{sec:pass_count}

We next test whether the benefit of the additional state grows with the number of iterations, holding state width fixed. With \(d=d_s=1024\), we train paired 410M Vanilla Looped and \method{} models using \(K=3,\ldots,7\) iterations. Each model uses 25B training tokens and is evaluated at the depth used during training. With identity mixing, \method{} has 411.10M parameters at every tested loop depth.

\begin{figure}[!ht]
\centering
\includegraphics[width=0.90\linewidth,trim=0 3bp 0 3bp,clip]{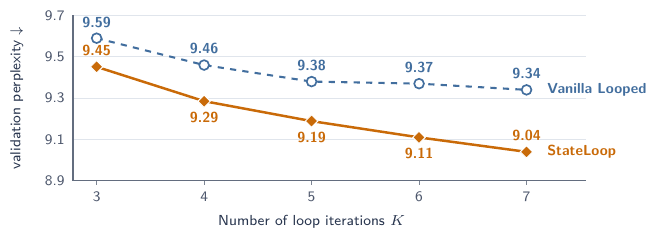}
\caption{\textbf{Loop depth and Pile PPL at 410M.} Models are trained for 25B tokens at each plotted \(K\). \method{} uses identity mixing and \(d=d_s=1024\). The PPL advantage over Vanilla Looped grows with trained loop depth.}
\label{fig:loop_count}
\end{figure}

Both models benefit from more iterations, with larger gains for \method{} (Figure~\ref{fig:loop_count}). Its PPL advantage over Vanilla Looped grows from approximately 0.14 at \(K=3\) to 0.30 at \(K=7\), consistent with persistent state becoming increasingly useful as recurrent depth increases.

\subsection{Decoding Efficiency}
\label{sec:decode_throughput}

Table~\ref{tab:decode_throughput} compares the three \method{} variants and wider-FFN baseline at \(K=4\) with Vanilla Looped at \(K=4\) and \(K=5\), measuring the cost of the state branch against an additional backbone iteration.

\begin{table}[!ht]
\caption{\textbf{Decoding efficiency at 1.4B.} Measurements use BF16, batch size 1, and a 1,024-token prefill followed by a 1,024-token continuation on one GPU. Latency is mean \(\pm\) standard deviation over five runs; changes are relative to Vanilla Looped at \(K=4\).}
\label{tab:decode_throughput}
\centering
\tablecaptiongap
\normalsize
\setlength{\tabcolsep}{3pt}
\renewcommand{\arraystretch}{1.00}
\begin{tabular*}{\linewidth}{@{\extracolsep{\fill}}lccccc@{}}
\toprule
Model & Params (M) & \(K\) & \shortstack{Latency\\(ms/token) $\downarrow$} & \shortstack{Throughput\\(tok/s) $\uparrow$} & \shortstack{Latency\\change} \\
\midrule
Vanilla Looped & 1420.92 & 4 & 121.224 $\pm$ 1.099 & 8.249 & baseline \\
Vanilla Looped & 1420.92 & 5 & 151.527 $\pm$ 1.064 & 6.599 & +24.997\% \\
Vanilla Looped + FFN & 1467.22 & 4 & 121.167 $\pm$ 1.293 & 8.253 & $-0.047\%$ \\
\midrule
\method{} & 1437.71 & 4 & 122.943 $\pm$ 1.483 & 8.134 & +1.418\% \\
\methodattnshort{} & 1454.49 & 4 & 123.778 $\pm$ 1.348 & 8.079 & +2.106\% \\
\methodmambashort{} & 1467.23 & 4 & 123.397 $\pm$ 0.923 & 8.104 & +1.792\% \\
\bottomrule
\end{tabular*}
\end{table}

During autoregressive decoding, \method{} inherits the per-loop backbone KV caches of Vanilla Looped. In addition, its identity-mixer variant maintains only a transient \(d_s\)-dimensional state for the current token across recurrent iterations. Attention and Mamba-2 mixers additionally maintain loop-specific token-axis caches.

The state branch adds only 1.42\%--2.11\% latency at \(K=4\), compared with 25.00\% for a fifth Vanilla Looped iteration. \method{} thus has 18.86\% lower latency than the \(K=5\) baseline.

\FloatBarrier
\section{Ablation and Analysis}
\label{sec:ablation_analysis}

We analyze 410M models at \(K=7\), using identity mixing for the \method{} variants. We first vary state width, then compare feedback designs, state persistence, and update rules. Interventions on a trained model examine how retained information and continued state updates contribute to prediction.

\subsection{State Width}
\label{sec:state_size}
\label{sec:depth_width}

\begin{figure}[!ht]
\centering
\includegraphics[width=\linewidth]{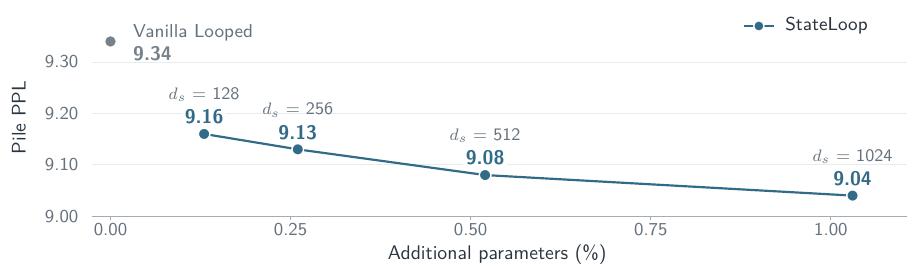}
\caption{\textbf{State width at 410M (\(K=7\), 25B tokens).} Pile PPL versus parameter overhead for identity mixing and backbone width \(d=1024\). Points are labeled with state width \(d_s\) and PPL; overhead is relative to the 406.90M-parameter Vanilla Looped baseline.}
\label{fig:state_width}
\end{figure}

We vary the persistent-state width \(d_s\in\{128,256,512,1024\}\) to test whether a smaller state preserves the language-modeling gains. Backbone width \(d=1024\) and loop depth \(K=7\) are fixed. \method{} uses the selective-SSM update with identity mixing. Figure~\ref{fig:state_width} reports PPL and parameter cost.

Halving the state width preserves most of the PPL gain while roughly halving the added parameter cost (Figure~\ref{fig:state_width}). Even the narrowest state improves over Vanilla Looped, showing that a compact persistent state is sufficient to provide gains.

\begin{table}[!ht]
\caption{\textbf{Persistent-state ablations on Pile at 410M.} All models use \(K=7\) and 25B training tokens. State variants use \(d_s=1024\) and identity mixing. State-only feedback uses the recurrent state as the next backbone input, whereas reset state sets the previous state to zero before each update. The GRU variant replaces the selective-SSM update while keeping feedback and final readout unchanged. Parameter counts are in millions.}
\label{tab:ablations}
\label{tab:pathway_control}
\label{tab:reset_state}
\label{tab:update_rule}
\centering
\tablecaptiongap
\normalsize
\setlength{\tabcolsep}{5pt}
\renewcommand{\arraystretch}{1.00}
\begin{tabular*}{0.82\linewidth}{@{\extracolsep{\fill}}llcc@{}}
\toprule
Ablation & Variant & Params (M) & PPL $\downarrow$ \\
\midrule
\multirow{2}{*}{Feedback} & State only & 409.00 & 9.44 \\
& \method{} & 411.10 & \textbf{9.04} \\
\midrule
\multirow{2}{*}{Persistence} & Reset state & 411.10 & 9.25 \\
& Persistent state & 411.10 & \textbf{9.04} \\
\midrule
\multirow{3}{*}{Update rule} & Vanilla Looped & 406.90 & 9.34 \\
& \method{} (GRU) & 416.34 & 9.09 \\
& \method{} & 411.10 & \textbf{9.04} \\
\bottomrule
\end{tabular*}
\end{table}

\subsection{Persistent-State Design}
\label{sec:core_ablation}
\label{sec:reset_state}
\label{sec:interface_control}

We examine which aspects of the persistent-state design contribute to these gains, considering feedback design, state persistence, and the update rule. All models are trained from scratch at 410M scale with \(K=7\) iterations and 25B training tokens. The \method{} variants use identity mixing.

\paragraph{Feedback design.}
The state-only alternative uses a gated state itself as the next backbone input \(Z^{k+1}\). From \(X^0=0\) and \(H^0=T(E)\), its recurrence for \(k=0,\ldots,K-2\) is
\begin{equation}
\begin{aligned}
X^{k+1}&=\operatorname{Cell}(H^k,X^k),\\
Z^{k+1}&=X^{k+1},\qquad H^{k+1}=T(Z^{k+1}).
\end{aligned}
\label{eq:state_as_carrier}
\end{equation}
Here \(X^k\in\mathbb{R}^{n\times d}\) has the backbone width, and \(\operatorname{Cell}\) applies the selective-SSM update in Equation~\ref{eq:ssm_vertical} at this width, with \(H^k\) as its input and \(X^k\) as its previous state. In \method{}, \(Z^{k+1}\) instead combines \(H^k\) with a projected read from the additional persistent state, as in Equation~\ref{eq:state_feedback}.

\method{} achieves lower PPL than state-only feedback
(Table~\ref{tab:ablations}). This result suggests an advantage for retaining the direct
connection between iterations and adding a state read, rather than replacing that path
with the recurrent state.

\paragraph{State persistence.}
We test persistence by setting the old state to zero before every update. The reset and persistent variants follow
\begin{equation}
\begin{aligned}
\text{Reset state}\quad&S^{k+1}=\eta^k\odot P^k,\\
\text{Persistent state}\quad&S^{k+1}=(1-\eta^k)\odot S^k+\eta^k\odot P^k.
\end{aligned}
\label{eq:reset_control}
\end{equation}

Resetting the state still improves over Vanilla Looped (9.25 vs.\ 9.34),
showing that the state branch within each iteration can improve performance even without
state persistence across iterations. Carrying the state across iterations yields a further
improvement to 9.04 under the same parameter count and training setup,
isolating an additional benefit from persistence
(Table~\ref{tab:ablations}).

\paragraph{State update rule.}
We retain the same projected input \(P^k=U^kW_u^{\top}\) and replace the selective-SSM update in Equation~\ref{eq:ssm_vertical} with a GRU \citep{cho2014gru}, keeping the feedback, final readout, and identity mixer unchanged. The GRU variant preserves most of the improvement over Vanilla Looped (Table~\ref{tab:ablations}), indicating that the benefit is not specific to the selective-SSM parameterization. The selective-SSM update achieves lower PPL with fewer parameters.

\subsection{Analysis of the Persistent State}
\label{sec:state_analysis}

We analyze a trained 410M \method{} with identity mixing and \(K=7\) on the full Pile validation set to examine how the persistent state supports later iterations and why continued state updates matter. All analyses share an unmodified baseline of 9.04 PPL.

\begin{figure}[!ht]
\centering
\includegraphics[width=0.96\linewidth,trim=0 3.5bp 0 1.5bp,clip]{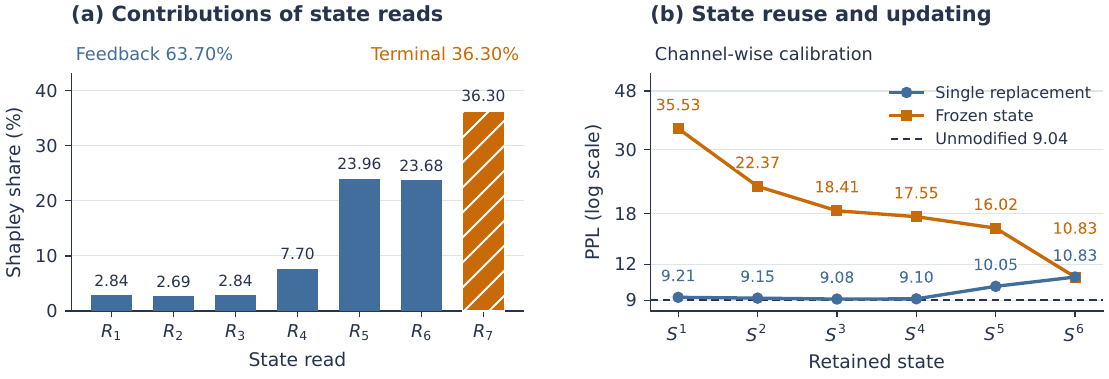}
\caption{\textbf{State analysis at 410M (\(K=7\), identity mixing).} Evaluated on the full Pile validation set. (a) Shapley shares of the NLL increase from masking reads. (b) Calibrated single replacement versus frozen-state reuse; the dashed line marks the unmodified model.}
\label{fig:state_analysis}
\end{figure}

\paragraph{State readouts.}
We mask a state read by setting its projected output to zero. To account for interactions, we evaluate all \(2^7\) masking combinations. Each read's exact Shapley value \citep{shapley1953value} averages the extra NLL increase from masking it across all removal orderings. Feedback reads account for 63.70\% of the attributed NLL increase, and the terminal read accounts for 36.30\% (Figure~\ref{fig:state_analysis}a). Contributions concentrate in \(R_5,R_6,R_7\), indicating greater reliance on the state in the final backbone transitions and prediction.

\paragraph{State replacement and freezing.}
To test whether earlier states remain usable at later boundaries, we first account for differences in channel scale and offset. Let \(\mu_{i,c}\) and \(\sigma_{i,c}\) be the mean and standard deviation of channel \(c\) at boundary \(i\), estimated from unmodified model evaluations on a separate calibration sample. To use \(S^i\) at boundary \(j\), we calibrate it as
\begin{equation}
\widehat S^{i\rightarrow j}_{t,c}
=\frac{\sigma_{j,c}}{\sigma_{i,c}+\epsilon}
\left(S^i_{t,c}-\mu_{i,c}\right)+\mu_{j,c}.
\label{eq:state_calibration}
\end{equation}
Here \(\epsilon=10^{-6}\) ensures numerical stability. Calibration applies the same scale and offset to all token positions \(t\) within each channel.

Single replacement reuses an earlier state once by replacing \(S^{i+1}\) with \(\widehat S^{i\rightarrow i+1}\), then resumes normal updates. Freezing reuses the same saved \(S^i\) at all remaining boundaries \(j>i\), recalibrating it to each target boundary. Both interventions continue running the backbone.

Single replacement causes small PPL increases at early boundaries but larger increases at \(S^6\) and \(S^7\), indicating that later updates become more important for prediction. Replacing \(S^4\) once with calibrated \(S^3\) yields 9.08 PPL, whereas freezing at \(S^3\) raises PPL to 18.41 (Figure~\ref{fig:state_analysis}b). This contrast suggests that earlier states remain partially usable at nearby boundaries, while continued state updates are important as the
backbone computation proceeds.

\FloatBarrier
\section{Conclusion}

\method{} equips each token with one additional persistent state to retain information across iterations. It reduces perplexity and improves average downstream accuracy over Vanilla Looped, with optional state mixers based on causal attention or Mamba-2 providing further downstream gains. The gains obtained with both selective-SSM and GRU updates support the persistent-state design. Its growing perplexity advantage over Vanilla Looped at greater trained depths further highlights the value of persistent state for deeper recurrent computation. State interventions show that earlier states remain partially usable at later boundaries, while continued state updates are important for prediction. These findings support explicit persistent state as a promising design choice for looped Transformers.

Our experiments cover models up to the 1.4B scale, with the controlled loop-depth analysis conducted at 410M. Extending this analysis to larger models can further characterize how persistent state interacts with recurrent depth.

\bibliography{paper_references}

\begin{thebibliography}{31}
\providecommand{\natexlab}[1]{#1}
\providecommand{\url}[1]{\texttt{#1}}
\expandafter\ifx\csname urlstyle\endcsname\relax
  \providecommand{\doi}[1]{doi: #1}\else
  \providecommand{\doi}{doi: \begingroup \urlstyle{rm}\Url}\fi

\bibitem[Bisk et~al.(2020)Bisk, Zellers, Le~Bras, Gao, and Choi]{bisk2020piqa}
Yonatan Bisk, Rowan Zellers, Ronan Le~Bras, Jianfeng Gao, and Yejin Choi.
\newblock {PIQA}: Reasoning about physical commonsense in natural language.
\newblock In \emph{Proceedings of the AAAI Conference on Artificial
  Intelligence}, volume~34, pages 7432--7439, 2020.
\newblock \doi{10.1609/aaai.v34i05.6239}.
\newblock URL \url{https://ojs.aaai.org/index.php/AAAI/article/view/6239}.

\bibitem[Bulatov et~al.(2022)Bulatov, Kuratov, and Burtsev]{bulatov2022rmt}
Aydar Bulatov, Yury Kuratov, and Mikhail Burtsev.
\newblock Recurrent memory transformer.
\newblock In \emph{Advances in Neural Information Processing Systems},
  volume~35, pages 11079--11091, 2022.
\newblock \doi{10.52202/068431-0805}.
\newblock URL
  \url{https://proceedings.neurips.cc/paper_files/paper/2022/hash/47e288629a6996a17ce50b90a056a0e1-Abstract-Conference.html}.

\bibitem[Cho et~al.(2014)Cho, van Merri{\"e}nboer, Gulcehre, Bahdanau,
  Bougares, Schwenk, and Bengio]{cho2014gru}
Kyunghyun Cho, Bart van Merri{\"e}nboer, Caglar Gulcehre, Dzmitry Bahdanau,
  Fethi Bougares, Holger Schwenk, and Yoshua Bengio.
\newblock Learning phrase representations using {RNN} encoder--decoder for
  statistical machine translation.
\newblock In \emph{Proceedings of the 2014 Conference on Empirical Methods in
  Natural Language Processing}, pages 1724--1734. Association for Computational
  Linguistics, 2014.
\newblock \doi{10.3115/v1/D14-1179}.
\newblock URL \url{https://aclanthology.org/D14-1179/}.

\bibitem[Clark et~al.(2018)Clark, Cowhey, Etzioni, Khot, Sabharwal, Schoenick,
  and Tafjord]{clark2018arc}
Peter Clark, Isaac Cowhey, Oren Etzioni, Tushar Khot, Ashish Sabharwal, Carissa
  Schoenick, and Oyvind Tafjord.
\newblock Think you have solved question answering? {Try} {ARC}, the {AI2}
  reasoning challenge.
\newblock \emph{arXiv preprint arXiv:1803.05457}, 2018.
\newblock URL \url{https://arxiv.org/abs/1803.05457}.

\bibitem[Dao and Gu(2024)]{dao2024transformers}
Tri Dao and Albert Gu.
\newblock Transformers are {SSM}s: Generalized models and efficient algorithms
  through structured state space duality.
\newblock In \emph{Proceedings of the 41st International Conference on Machine
  Learning}, volume 235 of \emph{Proceedings of Machine Learning Research},
  pages 10041--10071. PMLR, 2024.
\newblock URL \url{https://proceedings.mlr.press/v235/dao24a.html}.

\bibitem[Dehghani et~al.(2019)Dehghani, Gouws, Vinyals, Uszkoreit, and
  Kaiser]{dehghani2019universal}
Mostafa Dehghani, Stephan Gouws, Oriol Vinyals, Jakob Uszkoreit, and Lukasz
  Kaiser.
\newblock Universal transformers.
\newblock In \emph{International Conference on Learning Representations}, 2019.
\newblock URL \url{https://iclr.cc/virtual/2019/poster/1068}.

\bibitem[Frey et~al.(2026)Frey, Shomali, Bashir, Berghaus, Koehler, and
  Ali]{frey2026adaptive}
Markus Frey, Behzad Shomali, Ali~Hamza Bashir, David Berghaus, Joachim Koehler,
  and Mehdi Ali.
\newblock Adaptive loops and memory in transformers: Think harder or know more?
\newblock In \emph{Latent \& Implicit Thinking Workshop at ICLR 2026}, 2026.
\newblock URL \url{https://arxiv.org/abs/2603.08391}.

\bibitem[Gao et~al.(2020)Gao, Biderman, Black, Golding, Hoppe, Foster, Phang,
  He, Thite, Nabeshima, Presser, and Leahy]{gao2020pile}
Leo Gao, Stella Biderman, Sid Black, Laurence Golding, Travis Hoppe, Charles
  Foster, Jason Phang, Horace He, Anish Thite, Noa Nabeshima, Shawn Presser,
  and Connor Leahy.
\newblock The pile: An {800GB} dataset of diverse text for language modeling.
\newblock \emph{arXiv preprint arXiv:2101.00027}, 2020.
\newblock URL \url{https://arxiv.org/abs/2101.00027}.

\bibitem[Goyal et~al.(2024)Goyal, Ji, Rawat, Menon, Kumar, and
  Nagarajan]{goyal2023pause}
Sachin Goyal, Ziwei Ji, Ankit~Singh Rawat, Aditya~Krishna Menon, Sanjiv Kumar,
  and Vaishnavh Nagarajan.
\newblock Think before you speak: Training language models with pause tokens.
\newblock In \emph{International Conference on Learning Representations}, 2024.
\newblock URL \url{https://openreview.net/forum?id=ph04CRkPdC}.

\bibitem[Gu and Dao(2024)]{gu2023mamba}
Albert Gu and Tri Dao.
\newblock {Mamba}: Linear-time sequence modeling with selective state spaces.
\newblock In \emph{First Conference on Language Modeling}, 2024.
\newblock URL \url{https://openreview.net/forum?id=tEYskw1VY2}.

\bibitem[Hochreiter and Schmidhuber(1997)]{hochreiter1997lstm}
Sepp Hochreiter and J{\"u}rgen Schmidhuber.
\newblock Long short-term memory.
\newblock \emph{Neural Computation}, 9\penalty0 (8):\penalty0 1735--1780, 1997.
\newblock \doi{10.1162/neco.1997.9.8.1735}.

\bibitem[Lai et~al.(2017)Lai, Xie, Liu, Yang, and Hovy]{lai2017race}
Guokun Lai, Qizhe Xie, Hanxiao Liu, Yiming Yang, and Eduard Hovy.
\newblock {RACE}: Large-scale {R}e{A}ding comprehension dataset from
  examinations.
\newblock In \emph{Proceedings of the 2017 Conference on Empirical Methods in
  Natural Language Processing}, pages 785--794. Association for Computational
  Linguistics, 2017.
\newblock \doi{10.18653/v1/D17-1082}.
\newblock URL \url{https://aclanthology.org/D17-1082/}.

\bibitem[Merity et~al.(2017)Merity, Xiong, Bradbury, and
  Socher]{merity2017pointer}
Stephen Merity, Caiming Xiong, James Bradbury, and Richard Socher.
\newblock Pointer sentinel mixture models.
\newblock In \emph{International Conference on Learning Representations}, 2017.
\newblock URL \url{https://openreview.net/forum?id=Byj72udxe}.

\bibitem[Pagliardini et~al.(2024)Pagliardini, Mohtashami, Fleuret, and
  Jaggi]{pagliardini2024denseformer}
Matteo Pagliardini, Amirkeivan Mohtashami, Francois Fleuret, and Martin Jaggi.
\newblock {DenseFormer}: Enhancing information flow in transformers via depth
  weighted averaging.
\newblock In \emph{Advances in Neural Information Processing Systems},
  volume~37, 2024.
\newblock \doi{10.52202/079017-4336}.
\newblock URL
  \url{https://papers.nips.cc/paper_files/paper/2024/hash/f67449c7ab72f441d3a713b046c6818c-Abstract-Conference.html}.

\bibitem[Paperno et~al.(2016)Paperno, Kruszewski, Lazaridou, Pham, Bernardi,
  Pezzelle, Baroni, Boleda, and Fern{\'a}ndez]{paperno2016lambada}
Denis Paperno, Germ{\'a}n Kruszewski, Angeliki Lazaridou, Ngoc~Quan Pham,
  Raffaella Bernardi, Sandro Pezzelle, Marco Baroni, Gemma Boleda, and Raquel
  Fern{\'a}ndez.
\newblock The {LAMBADA} dataset: Word prediction requiring a broad discourse
  context.
\newblock In \emph{Proceedings of the 54th Annual Meeting of the Association
  for Computational Linguistics (Volume 1: Long Papers)}, pages 1525--1534.
  Association for Computational Linguistics, 2016.
\newblock \doi{10.18653/v1/P16-1144}.
\newblock URL \url{https://aclanthology.org/P16-1144/}.

\bibitem[Prairie et~al.(2026)Prairie, Novack, Berg-Kirkpatrick, and
  Fu]{prairie2026parcae}
Hayden Prairie, Zachary Novack, Taylor Berg-Kirkpatrick, and Daniel~Y. Fu.
\newblock Parcae: Scaling laws for stable looped language models.
\newblock \emph{arXiv preprint arXiv:2604.12946}, 2026.
\newblock URL \url{https://arxiv.org/abs/2604.12946}.

\bibitem[Radford et~al.(2019)Radford, Wu, Child, Luan, Amodei, and
  Sutskever]{radford2019language}
Alec Radford, Jeffrey Wu, Rewon Child, David Luan, Dario Amodei, and Ilya
  Sutskever.
\newblock Language models are unsupervised multitask learners.
\newblock Technical report, OpenAI, 2019.
\newblock URL
  \url{https://cdn.openai.com/better-language-models/language-models.pdf}.

\bibitem[Sakaguchi et~al.(2020)Sakaguchi, Le~Bras, Bhagavatula, and
  Choi]{sakaguchi2020winogrande}
Keisuke Sakaguchi, Ronan Le~Bras, Chandra Bhagavatula, and Yejin Choi.
\newblock {WinoGrande}: An adversarial {Winograd} schema challenge at scale.
\newblock In \emph{Proceedings of the AAAI Conference on Artificial
  Intelligence}, volume~34, pages 8732--8740, 2020.
\newblock \doi{10.1609/aaai.v34i05.6399}.
\newblock URL \url{https://ojs.aaai.org/index.php/AAAI/article/view/6399}.

\bibitem[Sapunov(2026)]{sapunov2026memory}
Grigory Sapunov.
\newblock Universal transformers need memory: Depth-state trade-offs in
  adaptive recursive reasoning.
\newblock \emph{arXiv preprint arXiv:2604.21999}, 2026.
\newblock URL \url{https://arxiv.org/abs/2604.21999}.

\bibitem[Saunshi et~al.(2025)Saunshi, Dikkala, Li, Kumar, and
  Reddi]{saunshi2025latent}
Nikunj Saunshi, Nishanth Dikkala, Zhiyuan Li, Sanjiv Kumar, and Sashank~J.
  Reddi.
\newblock Reasoning with latent thoughts: On the power of looped transformers.
\newblock In \emph{International Conference on Learning Representations}, 2025.
\newblock URL \url{https://openreview.net/forum?id=din0lGfZFd}.

\bibitem[Shapley(1953)]{shapley1953value}
Lloyd~S. Shapley.
\newblock A value for n-person games.
\newblock In Harold~W. Kuhn and Albert~W. Tucker, editors, \emph{Contributions
  to the Theory of Games II}, pages 307--317. Princeton University Press,
  Princeton, NJ, 1953.

\bibitem[Touvron et~al.(2023)Touvron, Lavril, Izacard, Martinet, Lachaux,
  Lacroix, Rozi{\`e}re, Goyal, Hambro, Azhar, Rodriguez, Joulin, Grave, and
  Lample]{touvron2023llama}
Hugo Touvron, Thibaut Lavril, Gautier Izacard, Xavier Martinet, Marie-Anne
  Lachaux, Timoth{\'e}e Lacroix, Baptiste Rozi{\`e}re, Naman Goyal, Eric
  Hambro, Faisal Azhar, Aur{\'e}lien Rodriguez, Armand Joulin, Edouard Grave,
  and Guillaume Lample.
\newblock {LLaMA}: Open and efficient foundation language models.
\newblock \emph{arXiv preprint arXiv:2302.13971}, 2023.
\newblock URL \url{https://arxiv.org/abs/2302.13971}.

\bibitem[Vaswani et~al.(2017)Vaswani, Shazeer, Parmar, Uszkoreit, Jones, Gomez,
  Kaiser, and Polosukhin]{vaswani2017attention}
Ashish Vaswani, Noam Shazeer, Niki Parmar, Jakob Uszkoreit, Llion Jones,
  Aidan~N. Gomez, Lukasz Kaiser, and Illia Polosukhin.
\newblock Attention is all you need.
\newblock In \emph{Advances in Neural Information Processing Systems},
  volume~30, pages 5998--6008, 2017.
\newblock URL
  \url{https://proceedings.neurips.cc/paper_files/paper/2017/hash/3f5ee243547dee91fbd053c1c4a845aa-Abstract.html}.

\bibitem[Welbl et~al.(2017)Welbl, Liu, and Gardner]{welbl2017sciq}
Johannes Welbl, Nelson~F. Liu, and Matt Gardner.
\newblock Crowdsourcing multiple choice science questions.
\newblock In \emph{Proceedings of the 3rd Workshop on Noisy User-generated
  Text}, pages 94--106. Association for Computational Linguistics, 2017.
\newblock \doi{10.18653/v1/W17-4413}.
\newblock URL \url{https://aclanthology.org/W17-4413/}.

\bibitem[Yang et~al.(2024)Yang, Lee, Nowak, and
  Papailiopoulos]{yang2023learning}
Liu Yang, Kangwook Lee, Robert~D. Nowak, and Dimitris Papailiopoulos.
\newblock Looped transformers are better at learning learning algorithms.
\newblock In \emph{International Conference on Learning Representations}, 2024.
\newblock URL \url{https://openreview.net/forum?id=HHbRxoDTxE}.

\bibitem[Yu et~al.(2026)Yu, Shu, Wang, Zhang, Wu, Li, Long, Chen, Xu, Su, and
  Zheng]{yu2026mesh}
Chengting Yu, Xiaobo Shu, Yadao Wang, Yizhen Zhang, Haoyi Wu, Jiaang Li, Rujiao
  Long, Ziheng Chen, Yuchi Xu, Wenbo Su, and Bo~Zheng.
\newblock {MeSH}: Memory-as-state-highways for recursive transformers.
\newblock In \emph{International Conference on Learning Representations}, 2026.
\newblock URL \url{https://openreview.net/forum?id=IhTrFvY7p3}.

\bibitem[Zeitoun et~al.(2026)Zeitoun, Torroba-Hennigen, and
  Kim]{zeitoun2026hyperloop}
Abbas Zeitoun, Lucas Torroba-Hennigen, and Yoon Kim.
\newblock Hyperloop transformers.
\newblock \emph{arXiv preprint arXiv:2604.21254}, 2026.
\newblock URL \url{https://arxiv.org/abs/2604.21254}.

\bibitem[Zellers et~al.(2019)Zellers, Holtzman, Bisk, Farhadi, and
  Choi]{zellers2019hellaswag}
Rowan Zellers, Ari Holtzman, Yonatan Bisk, Ali Farhadi, and Yejin Choi.
\newblock {HellaSwag}: Can a machine really finish your sentence?
\newblock In \emph{Proceedings of the 57th Annual Meeting of the Association
  for Computational Linguistics}, pages 4791--4800. Association for
  Computational Linguistics, 2019.
\newblock \doi{10.18653/v1/P19-1472}.
\newblock URL \url{https://aclanthology.org/P19-1472/}.

\bibitem[Zeng et~al.(2026)Zeng, Song, Huang, Wang, Li, He, Wang, Li, and
  Lin]{zeng2025ponder}
Boyi Zeng, Shixiang Song, Siyuan Huang, Yixuan Wang, He~Li, Ziwei He, Xinbing
  Wang, Zhiyu Li, and Zhouhan Lin.
\newblock {PonderLM}: Pretraining language models to ponder in continuous
  space.
\newblock In \emph{International Conference on Learning Representations}, 2026.
\newblock URL \url{https://openreview.net/forum?id=UrM4MNRYZm}.

\bibitem[Zhang and Sennrich(2019)]{zhang2019rmsnorm}
Biao Zhang and Rico Sennrich.
\newblock Root mean square layer normalization.
\newblock In \emph{Advances in Neural Information Processing Systems},
  volume~32, 2019.
\newblock URL
  \url{https://proceedings.neurips.cc/paper/2019/hash/1e8a19426224ca89e83cef47f1e7f53b-Abstract.html}.

\bibitem[Zhu et~al.(2025)Zhu, Wang, Hua, Zhang, Li, Que, Wei, Wen, Yin, Xing,
  Li, Shi, Ma, Li, Kergan, Smith, Qu, Hui, Wu, Min, Huang, Zhou, Ye, Liu, Yang,
  Shi, Lin, Zhao, Cai, Zhang, Huang, Bengio, and Eshraghian]{zhu2025ouro}
Rui-Jie Zhu, Zixuan Wang, Kai Hua, Tianyu Zhang, Ziniu Li, Haoran Que, Boyi
  Wei, Zixin Wen, Fan Yin, He~Xing, Lu~Li, Jiajun Shi, Kaijing Ma, Shanda Li,
  Taylor Kergan, Andrew Smith, Xingwei Qu, Mude Hui, Bohong Wu, Qiyang Min,
  Hongzhi Huang, Xun Zhou, Wei Ye, Jiaheng Liu, Jian Yang, Yunfeng Shi,
  Chenghua Lin, Enduo Zhao, Tianle Cai, Ge~Zhang, Wenhao Huang, Yoshua Bengio,
  and Jason Eshraghian.
\newblock Scaling latent reasoning via looped language models.
\newblock \emph{arXiv preprint arXiv:2510.25741}, 2025.
\newblock URL \url{https://arxiv.org/abs/2510.25741}.

\end{thebibliography}
\bibliographystyle{plainnat}

\end{document}